\pdfoutput=1
\documentclass[10pt,twocolumn,letterpaper]{article}

\usepackage{cvpr}
\usepackage[accsupp]{axessibility}

\usepackage{graphicx}
\usepackage{amsmath}
\usepackage{amssymb}
\usepackage{booktabs}
\usepackage{float}
\usepackage{cuted}
\usepackage{capt-of}
\usepackage{multirow}
\usepackage{bbding}

\usepackage[pagebackref,breaklinks,colorlinks]{hyperref}

\usepackage[capitalize]{cleveref}

\crefname{section}{Sec.}{Secs.}
\Crefname{section}{Section}{Sections}
\Crefname{table}{Table}{Tables}
\crefname{table}{Tab.}{Tabs.}

\def\etal{{\em et al.}}

\newcommand{\sota}{state-of-the-art}

\newcommand{\backwardwarp}{\overleftarrow{\mathcal{W}}}
\newcommand*\samethanks[1][\value{footnote}]{\footnotemark[#1]}

\def\cvprPaperID{1448}
\def\confName{CVPR}
\def\confYear{2023}

\begin{document}
\title{A Dynamic Multi-Scale Voxel Flow Network for Video Prediction}
\author{Xiaotao Hu$^{1,2}$
		~~~~
		Zhewei Huang$^{2}$
        ~~~~
		Ailin Huang$^{2,3}$
        ~~~~
            Jun Xu$^{4,}$\thanks{Corresponding authors.}
        ~~~~
            Shuchang Zhou$^{2,}$\samethanks{}\\
		$^{1}$College of Computer Science, Nankai University~~~~$^{2}$Megvii Technology\\ $^{3}$Wuhan University~~~~
        $^{4}$School of Statistics and Data Science, Nankai University\\
		{\tt\small \{huxiaotao, huangzhewei, huangailin, zhoushuchang\}@megvii.com}\tt\small, nankaimathxujun@gmail.com\\
  \url{https://huxiaotaostasy.github.io/DMVFN/}
	}
	\maketitle
 
\begin{abstract}
The performance of video prediction has been greatly boosted by advanced deep neural networks.
However, most of the current methods suffer from large model sizes and require extra inputs, \eg, semantic/depth maps, for promising performance.
For efficiency consideration, in this paper, we propose a Dynamic Multi-scale Voxel Flow Network (DMVFN) to achieve better video prediction performance at lower computational costs with only RGB images, than previous methods.
The core of our DMVFN is a differentiable routing module that can effectively perceive the motion scales of video frames.
Once trained, our DMVFN selects adaptive sub-networks for different inputs at the inference stage.
Experiments on several benchmarks demonstrate that our DMVFN is an order of magnitude faster than Deep Voxel Flow~\cite{dvf} and surpasses the state-of-the-art iterative-based OPT~\cite{wu2022optimizing} on generated image quality.
	\end{abstract}
	\section{Introduction}

Video prediction aims to predict future video frames from the current ones. The task potentially benefits the study on representation learning~\cite{oprea2020review} and downstream forecasting tasks such as human motion prediction~\cite{martinez2017human}, autonomous driving~\cite{castrejon2019improved}, and climate change~\cite{shi2015convolutional}, \etc. During the last decade, video prediction has been increasingly studied in both academia and industry community~\cite{chandra2019traphic,byeon2018contextvp}.

\begin{figure}[tb]
	\centering
	\includegraphics[width=0.47\textwidth]{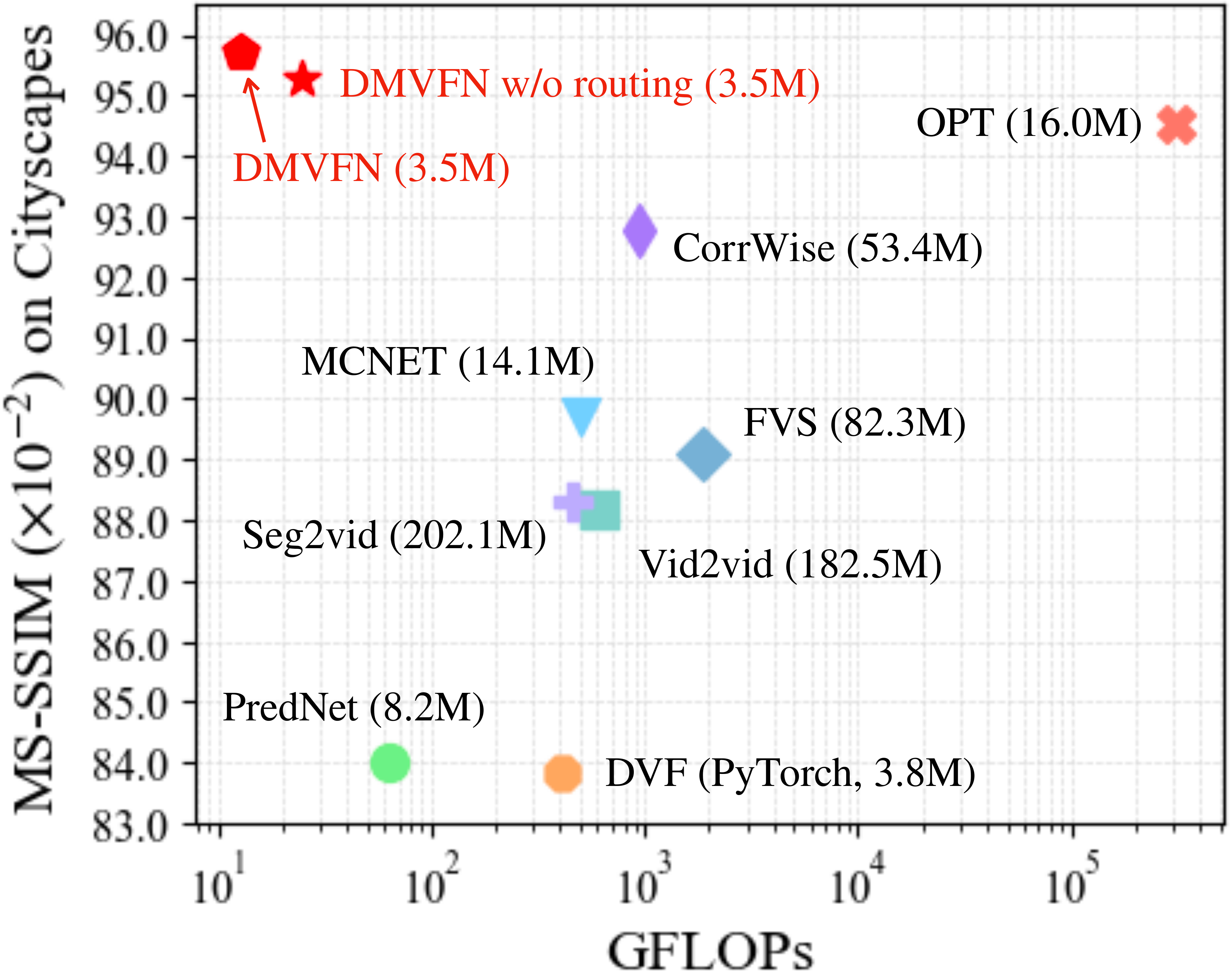}
	\caption{
\textbf{Average MS-SSIM and GFLOPs of different video prediction methods} on Cityscapes~\cite{cityscapes}. The parameter amounts are provided in brackets. DMVFN outperforms previous methods in terms of image quality, parameter amount, and GFLOPs.
}
	\label{fig:performance}
\end{figure}

Video prediction is challenging because of the diverse and complex motion patterns in the wild, in which accurate motion estimation plays a crucial role~\cite{prednet,mcnet,dvf}.
Early methods~\cite{prednet,mcnet} along this direction mainly utilize recurrent neural networks~\cite{hochreiter1997long} to capture temporal motion information for video prediction. To achieve robust long-term prediction, the works of~\cite{seg2vid,vid2vid,fvs} additionally exploit the semantic or instance segmentation maps of video frames for semantically coherent motion estimation in complex scenes. However, the semantic or instance maps may not always be available in practical scenarios, which limits the application scope of these video prediction methods~\cite{seg2vid,vid2vid,fvs}. To improve the prediction capability while avoiding extra inputs, the method of OPT~\cite{wu2022optimizing} utilizes only RGB images to estimate the optical flow of video motions in an optimization manner with impressive performance. However, its inference speed is largely bogged down mainly by the computational costs of pre-trained optical flow model~\cite{raft} and frame interpolation model~\cite{rife} used in the iterative generation.

The motions of different objects between two adjacent frames are usually of different scales. This is especially evident in high-resolution videos with meticulous details~\cite{xvfi}. The spatial resolution is also of huge differences in real-world video prediction applications. To this end, it is essential yet challenging to develop a single model for multi-scale motion estimation. An early attempt is to extract multi-scale motion cues in different receptive fields by employing the encoder-decoder architecture~\cite{dvf}, but in practice it is not flexible enough to deal with complex motions.

In this paper, we propose a Dynamic Multi-scale Voxel Flow Network (DMVFN) to explicitly model the complex motion cues of diverse scales between adjacent video frames by dynamic optical flow estimation. Our DMVFN is consisted of several Multi-scale Voxel Flow Blocks (MVFBs), which are stacked in a sequential manner. On top of MVFBs, a light-weight Routing Module is proposed to adaptively generate a routing vector according to the input frames, and to dynamically select a sub-network for efficient future frame prediction. We conduct experiments on four benchmark datasets, including Cityscapes~\cite{cityscapes}, KITTI~\cite{kitti}, DAVIS17~\cite{davis}, and Vimeo-Test~\cite{vimeo}, to demonstrate the comprehensive advantages of our DMVFN over representative video prediction methods in terms of visual quality, parameter amount, and computational efficiency measured by floating point operations~(FLOPs). A glimpse of comparison results by different methods is provided in Figure~\ref{fig:performance}. One can see that our DMVFN achieves much better performance in terms of accuracy and efficiency on the Cityscapes~\cite{cityscapes} dataset. Extensive ablation studies validate the effectiveness of the components in our DMVFN for video prediction.

In summary, our contributions are mainly three-fold:
\begin{itemize}
    \item We design a light-weight DMVFN to accurately predict future frames with only RGB frames as inputs. Our DMVFN is consisted of new MVFB blocks that can model different motion scales in real-world videos.
    
    \item We propose an effective Routing Module to dynamically select a suitable sub-network according to the input frames. The proposed Routing Module is end-to-end trained along with our main network DMVFN.
    
    \item Experiments on four benchmarks show that our DMVFN achieves state-of-the-art results while being an order of magnitude faster than previous methods.
\end{itemize}
	
\section{Related Work}

\subsection{Video Prediction}
Early video prediction methods~\cite{prednet,mcnet,dvf} only utilize RGB frames as inputs. For example, PredNet~\cite{prednet} learns an unsupervised neural network, with each layer making local predictions and forwarding deviations from those predictions to subsequent network layers. MCNet~\cite{mcnet} decomposes the input frames into motion and content components, which are processed by two separate encoders. DVF~\cite{dvf} is a fully-convolutional encoder-decoder network synthesizing intermediate and future frames by approximating voxel flow for motion estimation. Later, extra information is exploited by video prediction methods in pursuit of better performance. For example, the methods of Vid2vid~\cite{vid2vid}, Seg2vid~\cite{seg2vid}, HVP~\cite{hvp}, and SADM~\cite{sadm} require additional semantic maps or human pose information for better video prediction results. Additionally, Qi~\etal~\cite{qi} used extra depth maps and semantic maps to explicitly inference scene dynamics in 3D space. FVS~\cite{fvs} separates the inputs into foreground objects and background areas by semantic and instance maps, and uses a spatial transformer to predict the motion of foreground objects. In this paper, we develop a light-weight and efficient video prediction network that requires only sRGB images as the inputs.

\subsection{Optical Flow}
Optical flow estimation aims to predict the per-pixel motion between adjacent frames. Deep learning-based optical flow methods~\cite{pwcnet,raft,jonschkowski2020matters,luo2021upflow,han2022realflow} have been considerably advanced ever since Flownet~\cite{flownet}, a pioneering work to learn optical flow network from synthetic data. Flownet2.0~\cite{flownet2} improves the accuracy of optical flow estimation by stacking sub-networks for iterative refinement. A coarse-to-fine spatial pyramid network is employed in SPynet~\cite{spynet} to estimate optical flow at multiple scales. PWC-Net~\cite{pwcnet} employs feature warping operation at different resolutions and uses a cost volume layer to refine the estimated flow at each resolution. RAFT~\cite{raft} is a lightweight recurrent network sharing weights during the iterative learning process. FlowFormer~\cite{flowformer} utilizes an encoder to output latent tokens and a recurrent decoder to decode features, while refining the estimated flow iteratively. In video synthesis, optical flow for downstream tasks~\cite{vimeo,dvf,rife,qvi,zhang2023extracting} is also a hot research topic. Based on these approaches, we aim to design a flow estimation network that can adaptively operate based on each sample for the video prediction task.

\begin{figure*}[ht]
	\centering
	\includegraphics[width=0.9\linewidth]{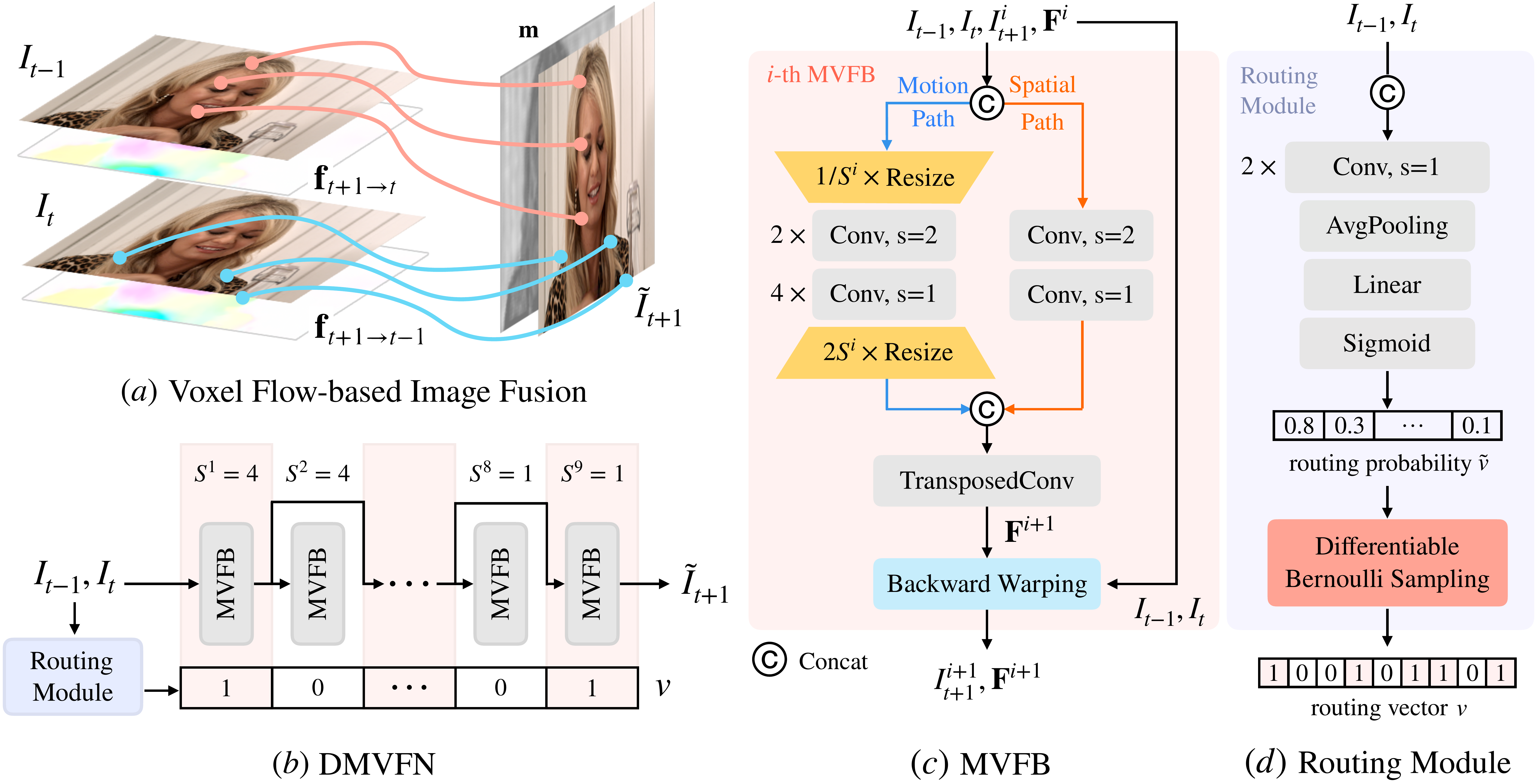}
	\caption{\textbf{Overview of the proposed Dynamic Multi-scale Voxel Flow Network (DMVFN)}.
$(a)$: To predict a future frame, we use the \textbf{voxel flow}~\cite{dvf} to guide the pixel fusion of the input frames. The voxel flow contains the prediction of object motion and occlusion. $(b)$: \textbf{DMVFN} contains several MVFBs with decreasing scaling factor $S^i$.
According to the routing vector $v$ estimated by a Routing Module, a sub-network is selected to process the input image. 
$(c)$: Each \textbf{MVFB} has a scaling factor $S^i$, which means that the motion path is performed on images whose sizes are $1/S^i$ of the original. $(d)$: Two consecutive frames are fed into several neural layers and a Differentiable Bernoulli sample to generate the hard routing vector.
}
	\label{fig:predict}
\end{figure*}

\subsection{Dynamic Network}
The design of dynamic networks is mainly divided into three categories: spatial-wise, temporal-wise, and sample-wise~\cite{dynamicnn}. Spatial-wise dynamic networks perform adaptive operations in different spatial regions to reduce computational redundancy with comparable performance~\cite{sbnet,dynamicconv,mga}. In addition to the spatial dimension, dynamic processing can also be applied in the temporal dimension. Temporal-wise dynamic networks~\cite{yeung2016end, su2016leaving, wu2019adaframe} improve the inference efficiency by performing less or no computation on unimportant sequence frames.
To handle the input in a data-driven manner, sample-wise dynamic networks adaptively adjust network structures to side-off the extra computation~\cite{skipnet,veit2018convolutional}, or adaptively change the network parameters to improve the performance~\cite{harley2017segmentation,su2019pixel,dcn,dcn2}. Designing and training a dynamic network is not trivial since it is difficult to directly enable a model with complex topology connections. We need to design a well-structured and robust model before considering its dynamic mechanism. In this paper, we propose a module to dynamically perceive the motion magnitude of input frames to select the network structure.

\section{Methodology}

\subsection{Background}
\label{sec:background}

\paragraph{Video prediction.}
Given a sequence of past $t$ frames $\{I_{i}\in \mathbb{R}^{h\times w\times 3}|i=1,...,t\}$, video prediction aims to predict the future frames $\{\tilde{I}_{t+1}, \tilde{I}_{t+2}, \tilde{I}_{t+3} ,\ ...\}$. The inputs of our video prediction model are only the two consecutive frames $I_{t-1}$ and $I_t$. We concentrate on predicting $\tilde{I}_{t+1}$, and iteratively predict future frames $\{\tilde{I}_{t+2}, \tilde{I}_{t+3} ,\ ...\}$ in a similar manner. Denote the video prediction model as $G_{\theta}(I_{t-1},I_{t})$, where $\theta$ is the set of model parameters to be learned, the learning objective is to minimize the difference between $\tilde{I}_{t+1}=G_{\theta}(I_{t-1},I_{t})$ and the ``ground truth'' $I_{t+1}$.

\paragraph{Voxel flow.} Considering the local consistency in space-time, the pixels of a generated future frame come from nearby regions of the previous frames~\cite{zhou2016view,vimeo}. In video prediction task, researchers estimate optical flow $\textbf{f}_{t+1 \rightarrow t}$ from $I_{t+1}$ to $I_t$~\cite{dvf}. And the corresponding frame is obtained using the pixel-wise backward warping~\cite{backwarp}~(denoted as $\backwardwarp$). In addition, to deal with the occlusion, some methods~\cite{superslomo,dvf} further introduce a fusion map $\textbf{m}$ to fuse the pixels of $I_{t}$ and $I_{t-1}$. The final predicted frame is obtained by the following formulation~(Figure~\ref{fig:predict} $(a)$):
\begin{equation}
\hat{I}_{t+1\leftarrow{t-1}} = \backwardwarp( I_{t-1},   \textbf{f}_{ t+1 \rightarrow t-1} ),
\end{equation}
\begin{equation}
     \hat{I}_{t+1 \leftarrow{t}} = \backwardwarp( I_t,  \textbf{f}_{t+1 \rightarrow t} ),
\end{equation}
\begin{equation}
     \tilde{I}_{t+1} = \hat{I}_{t+1\leftarrow{t-1}} \times \textbf{m} + \hat{I}_{t+1\leftarrow{t}} \times (1-\textbf{m}).
\end{equation}
Here, $\hat{I}_{t+1 \leftarrow{t}}$ and $\hat{I}_{t+1 \leftarrow{t-1}}$ are intermediate warped images. To simplify notations, we refer to the optical flows $\textbf{f}_{t+1 \rightarrow t},\textbf{f}_{t+1 \rightarrow t-1}$ and the fusion map $\textbf{m}$ collectively as the voxel flow $\textbf{F}_{t+1}$, similar to the notations in~\cite{dvf}. The above equations can be simplified to the following form:
\begin{equation}
     \tilde{I}_{t+1} = \backwardwarp( I_{t-1}, I_t,  \textbf{F}_{t+1} ).
\end{equation}

\begin{figure*}[htbp]
	\centering
	\includegraphics[width=0.95\linewidth]{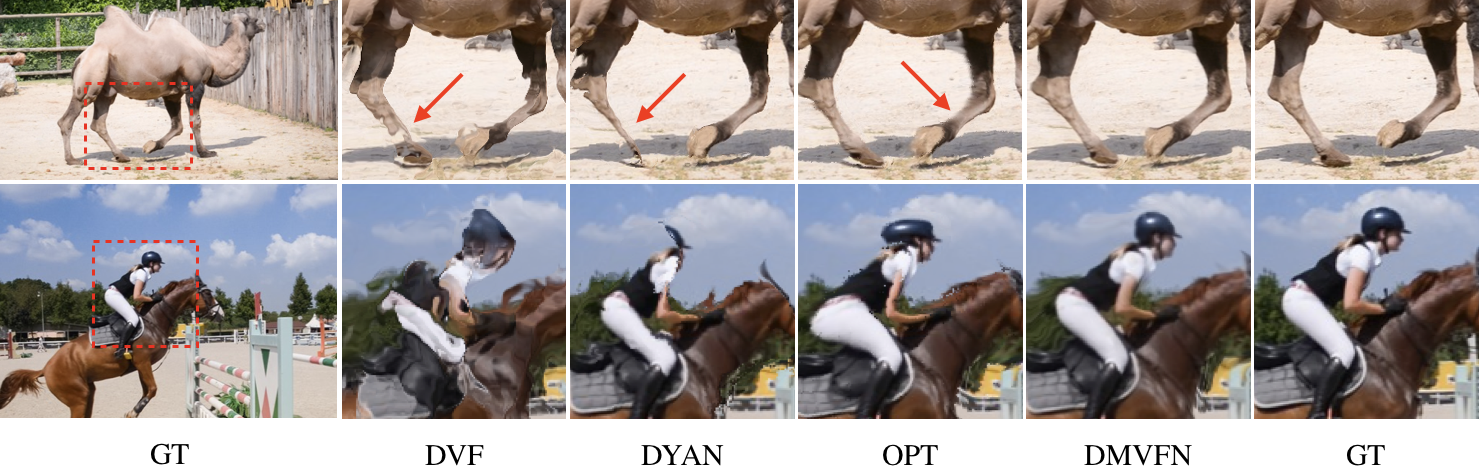}
	\caption{\textbf{Visual comparison of ($t+1$)-th frame predicted from $t$-th and ($t-1$)-th frames on the DAVIS17-Val~\cite{davis}}.}
	\label{fig:davis17}
 \vspace{-1em}
\end{figure*}

\subsection{Dynamic Multi-Scale Voxel Flow Network}
\label{sec:dmvfn}

\paragraph{MVFB.} To estimate the voxel flow, DVF~\cite{dvf} assumes that all optical flows are locally linear and temporally symmetric around the targeted time, which may be unreasonable for large-scale motions. To address the object position changing issue~\cite{rife} in adjacent frames, OPT~\cite{wu2022optimizing} uses flow reversal layer~\cite{qvi} to convert forward flows to backward flows. We aim to estimate voxel flow end-to-end without introducing new components and unreasonable constraints.

We denote the $i$-th MVFB as $f^i_{MVFB}(\cdot)$. It learns to approximate target voxel flow $\textbf{F}^{i}_{t+1}$ by
taking two frames $I_{t-1}$ and $I_t$, the synthesized frame $\tilde{I}^{i-1}_{t+1}$, and the voxel flow estimated by previous blocks $\textbf{F}^{i-1}_{t+1}$ as inputs.

The architecture of our MVFB is shown in Figure~\ref{fig:predict} $(c)$. To capture the large motion while retaining the original spatial information, we construct a two-branch network structure~\cite{yu2018bisenet}. This design inherits from pyramidal optical flow estimation~\cite{pwcnet,spynet}. In the \textbf{motion path}, the input is down-sampled by a scaling factor $S^i$ to facilitate the expansion of the receptive field. Another \textbf{spatial path} operates at high resolution to complement the spatial information. We denote $\tilde{I}^{i}_{t+1} $ as the output of the $i$-th MVFB. Formally,
\begin{equation}
\tilde{I}^{i}_{t+1}, \textbf{F}^i_{t+1} = f_{\mathrm{MVFB}}^i (I_{t-1}, I_t, \tilde{I}^{i-1}_{t+1}, \textbf{F}^{i-1}_{t+1}, S^i).
\end{equation}
The initial values of $\tilde{I}_{t+1}^{0}$ and $\textbf{F}^{0}_{t+1}$ are set to zero. As illustrated in Figure~\ref{fig:predict}~$(b)$, our DMVFN contains $9$ MVFBs. To generate a future frame, we iteratively refine a voxel flow~\cite{dvf} and fuse the pixels of the input frames. 

Many optical flow estimation methods predict the flow field on a small image, and then refine it on a large image~\cite{pwcnet,gmflow}. For simplicity and intuition, we consider decreasing scaling factor sequences. Finally, the scaling factors is experimentally set as $[4,4,4,2,2,2,1,1,1]$.

%

\paragraph{DMVFN.} Different pairs of adjacent frames have diverse motion scales and different computational demands. An intuitive idea is to adaptively select dynamic architectures conditioned on each input. We then perform dynamic routing within the~\textbf{super network} (the whole architecture)~\cite{dynamicnn}, including multiple possible paths. DMVFN saves redundant computation for samples with small-scale motion and preserves the representation ability for large-scale motion.

To make our DMVFN end-to-end trainable, we design a differentiable Routing Module containing a tiny neural network to estimate routing vector $v$ for each input sample. Based on this vector, our DMVFN dynamically selects a sub-network to process the input data. As the figure shows, some blocks are skipped during inference.

Different from some dynamic network methods that can only continuously select the first several blocks ($n$ options)~\cite{branchynet,bolukbasi2017adaptive}, DMVFN is able to choose paths freely ($2^{n}$ options). DMVFN trains different sub-networks in the \textbf{super network} with various possible inference paths and uses dynamic routing inside the \textbf{super network} during inference to reduce redundant computation while maintaining the performance. A dynamic routing vector $v \in { \{ 0,1 \} }^{n}$ is predicted by the proposed Routing Module. For the $i$-th MVFN block of DMVFN, we denote $v_i$ as the reference of whether processing the reached voxel flow $\textbf{F}^{i-1}_{t+1}$ and the reached predicted frame $\tilde{I}^{i-1}_{t+1}$. The path $f^i_\mathrm{MVFB}$ to the $i$-th block from the last block will be activated only when $v_i=1$. Formally, 
\begin{equation}
\label{eqn:dualbranchselection}
\tilde{I}^{i}_{t+1},\textbf{F}^{i}_{t+1}=
\begin{cases}
f^i_\mathrm{MVFB}(\tilde{I}^{i-1}_{t+1},\textbf{F}^{i-1}_{t+1}), & v_i=1 \\
\\
\tilde{I}^{i-1}_{t+1},\textbf{F}^{i-1}_{t+1}, & v_i=0.
\end{cases}
\end{equation}
During the training phase, to enable the backpropagation of Eqn.~(\ref{eqn:dualbranchselection}), we use $v_i$ and $(1-v_i)$ as the weights of the two branches and average their outputs.

\begin{table*}[th]
\centering
\caption{
\textbf{Quantitative results of different methods on the Cityscapes~\cite{cityscapes}, and KITTI~\cite{kitti} datasets}. ``RGB", ``F", ``S" and ``I" denote the video frames, optical flow, semantic map, and instance map, respectively. We denote our DMVFN without routing module as ``DMVFN~(w/o r)". FVS~\cite{fvs} integrates a segmentation model~\cite{zhu2019improving} on KITTI~\cite{kitti} to obtain the semantic maps. ``N/A'' means not available. }
\label{table:cityscapes}
\resizebox{0.96\linewidth}{!}{
\begin{tabular}{rccccccccccccccc}
\toprule
\multirow{3}{*}[-0.28em]{Method}& \multirow{3}{*}[-0.28em]{Inputs} &  \multicolumn{7}{c}{Cityscapes-Train$\rightarrow$Cityscapes-Test~\cite{cityscapes}} & \multicolumn{7}{c}{KITTI-Train$\rightarrow$KITTI-Test~\cite{kitti}} \\
\cmidrule(l{7pt}r{7pt}){3-9} \cmidrule(l{7pt}r{7pt}){10-16}
& & \multirow{2}{*}[-0.14em]{GFLOPs} & \multicolumn{3}{c}{MS-SSIM ($\times10^{-2}$) $\uparrow$} & \multicolumn{3}{c}{LPIPS ($\times10^{-2}$) $\downarrow$} &  \multirow{2}{*}[-0.14em]{GFLOPs}  & \multicolumn{3}{c}{MS-SSIM ($\times10^{-2}$) $\uparrow$} & 
\multicolumn{3}{c}{LPIPS ($\times10^{-2}$) $\downarrow$}\\
&  &  & $t+1$ & $t+3$ & $t+5$  & $t+1$  & $t+3$ & $t+5$ & & $t+1$ & $t+3$ & $t+5$  & $t+1$  & $t+3$ & $t+5$  \\ 
\midrule
\midrule
Vid2vid~\cite{vid2vid} & RGB+S &   603.79    & 88.16 & 80.55&75.13 & 10.58 & 15.92& 20.14 & N/A & N/A & N/A& N/A & N/A & N/A& N/A  \\
Seg2vid~\cite{seg2vid} & RGB+S &  455.84  & 88.32 & N/A & 61.63 & 9.69 &  N/A& 25.99& N/A & N/A & N/A& N/A & N/A& N/A& N/A\\ 
FVS~\cite{fvs}   & RGB+S+I&  1891.65 & 89.10 & 81.13 & 75.68 & 8.50 & \textbf{12.98} & 16.50 & \textbf{768.96} & 79.28 & 67.65& 60.77  & 18.48 & 24.61& 30.49\\
SADM~\cite{sadm}   & RGB+S+F&  N/A & \textbf{95.99} & N/A & \textbf{83.51} & \textbf{7.67} & N/A & \textbf{14.93} & N/A & \textbf{83.06} & \textbf{72.44} & \textbf{64.72}  & \textbf{14.41} & \textbf{24.58} &  \textbf{31.16}\\

\midrule

PredNet~\cite{prednet} & RGB & 62.62 & 84.03 & 79.25& 75.21 & 25.99 & 29.99& 36.03 & 25.44 & 56.26 &  51.47& 47.56 & 55.35 & 58.66& 62.95               \\
MCNET~\cite{mcnet} & RGB & 502.80 & 89.69 & 78.07& 70.58 & 18.88 & 31.34& 37.34 & 204.26 & 75.35  & 63.52& 55.48 & 24.05  & 31.71& 37.39  \\
DVF~\cite{dvf} & RGB & 409.78 & 83.85 & 76.23& 71.11 & 17.37 & 24.05& 28.79& 166.47 & 53.93 & 46.99& 42.62  & 32.47  & 37.43& 41.59         \\
CorrWise~\cite{corrwise} & RGB & 944.29 & 92.80 & N/A & \textbf{83.90} & 8.50 & N/A & 15.00 & 383.62 & 82.00 & N/A & 66.70 & 17.20 & N/A & \textbf{25.90}\\ 
OPT~\cite{wu2022optimizing} & RGB &  313482.15  & 94.54 & 86.89 & 80.40 & 6.46 & 12.50&17.83 & 127431.71 & 82.71 & 69.50 & 61.09 & 12.34 & 20.29 & 26.35\\

DMVFN~(w/o r)  & RGB &  24.51  & 95.29 & 87.91 & 81.48 & 5.60 & 10.48 & 14.91 & 9.96 &88.06 & 76.53 & 68.29 & \textbf{10.70} & 19.28 & 26.13 \\
DMVFN  & RGB & \textbf{12.71} & \textbf{95.73} & \textbf{89.24}& 83.45 & \textbf{5.58} & \textbf{10.47}& \textbf{14.82} & \textbf{5.15} & \textbf{88.53} & \textbf{78.01} & \textbf{70.52} & 10.74 & \textbf{19.27} & 26.05    \\

\midrule
\end{tabular}
}
\end{table*}

In the iterative scheme of our DMVFN, each MVFB essentially refines the current voxel flow estimation to a new one. This special property allows our DMVFN to skip some MVFBs for every pair of input frames. Here, we design a differentiable and efficient routing module for learning to trade-off each MVFB block. This is achieved by predicting a routing vector $v\in\{0,1\}^{n}$ to identify the proper sub-network (\eg, 0 for deactivated MVFBs, 1 for activated MVFBs). We implement the routing module by a small neural network~($\sim1/6$ GFLOPs of the \textbf{super network}), and show its architecture in Figure~\ref{fig:predict}~$(d)$. It learns to predict the probability $\tilde{v}$ of choosing MVFBs by:
\begin{equation}
\tilde{v} = \text{Linear}(\text{AvgPooling}(\text{Convs}(I_{t-1},I_t))), 
\end{equation}
\begin{equation}
v = \text{Bernoulli-Sampling}(\tilde{v}). 
\end{equation}

\paragraph{Differentiable Routing.} To train the proposed Routing Module, we need to constrain the probability values to prevent the model from falling into trivial solutions (\eg, select all blocks). On the other hand, we allow this module to participate in the gradient calculation to achieve end-to-end training. We introduce the Gumbel Softmax~\cite{gumbel} and the Straight-Through Estimator (STE)~\cite{bengio2013estimating} to tackle this issue.

One popular method to make the routing probability $\tilde{v}$ learnable is the \textbf{Gumbel Softmax} technique~\cite{gumbel,channel}. By treating the selection of each MVFB as a binary classification task, the soft dynamic routing vector $v\in\mathbb{R}^{n}$ is
\begin{equation}
v_i = \frac{\exp\big (\frac{1}{\tau}(\tilde{v}_{i} + G_{i})\big )}{\exp\big (\frac{1}{\tau}( \tilde{v}_{i} + G_{i})                   \big )+\exp\big (\frac{1}{\tau}(2-\tilde{v}_{i}-G_{i})\big )},
\end{equation}
where $i=1,...,n$, $G_{i}\in\mathbb{R}$ is Gumbel noise following the Gumbel$(0,1)$ distribution, and $\tau$ is a temperature parameter. We start at a very high temperature to ensure that all possible paths become candidates, and then the temperature is attenuated to a small value to approximate one-hot distribution. To encourage the sum of the routing vectors $\{v_{i}\}_{i=1}^{n}$ to be small, we add the regularization term ($\frac{1}{n}\sum^{n}_{i=1}v_i$) to the final loss function. However, we experimentally find that our DMVFN usually converges to an input-independent structure when temperature decreases. We conjecture that the control of the temperature parameter $\tau$ and the design of the regularization term require further study.

Inspired by previous research on low-bit width neural networks~\cite{dorefanet, hubara2017quantized}, we adopt STE for Bernoulli Sampling~(\textbf{STEBS}) to make the binary dynamic routing vector differentiable. An STE can be regarded as an operator that has arbitrary forward and backward operations. Formally,
\begin{equation}
\tilde{w}_i = \min(\beta\times n\times\sigma(\tilde{v}_i) / \sum_i^n \sigma(\tilde{v}_i), 1),
\label{normalize}
\end{equation}
\begin{equation}
\textbf{STE Forward}: v_i \sim \text{Bernoulli}(\tilde{w}_i),
\end{equation}
\begin{equation}
\textbf{STE Backward}: \frac{\partial o}{\partial \tilde{w}} = \frac{\partial o}{\partial v},
\end{equation}
where $\sigma$ is the Sigmoid function and we denote the objective function as $o$. We use the well-defined gradient $\frac{\partial o}{\partial v}$ as an approximation for $\frac{\partial o}{\partial \tilde{w}}$ to construct the backward pass. In Eqn.~(\ref{normalize}), we normalize the sample rate. During training, $\beta$ is fixed at $0.5$. We can adjust the hyper-parameter $\beta$ to control the complexity in the inference phase.

\begin{table*}[th]
\centering
\caption{
\textbf{Quantitative results on the DAVIS17-Val~\cite{davis} and Vimeo90K-Test~\cite{vimeo} benchmarks}. We denote DMVFN without routing as ``DMVFN~(w/o r)''. ``N/A'' means not available. }
\centering
\resizebox{0.95\linewidth}{!}{
\begin{tabular}{rcccccccc}
\toprule
\multirow{3}{*}[-0.28em]{Method}& \multicolumn{5}{c}{UCF101-Train$\rightarrow$DAVIS17-Val} &  \multicolumn{3}{c}{UCF101-Train$\rightarrow$Vimeo90K-Test} \\
\cmidrule(l{7pt}r{7pt}){2-6} \cmidrule(l{7pt}r{7pt}){7-9}
& \multirow{2}{*}[-0.14em]{GFLOPs  $\downarrow$} & \multicolumn{2}{c}{MS-SSIM ($\times10^{-2}$) $\uparrow$} & 
\multicolumn{2}{c}{LPIPS ($\times10^{-2}$) $\downarrow$} &  \multirow{2}{*}[-0.14em]{GFLOPs  $\downarrow$}  & 
\multicolumn{1}{c}{MS-SSIM ($\times10^{-2}$) $\uparrow$} & 
\multicolumn{1}{c}{LPIPS ($\times10^{-2}$)  $\downarrow$} \\
&  &$t+1$   & $t+3$ & $t+1$   & $t+3$    &  & $t+1$  & $t+1$ \\
\midrule
\midrule
DVF~\cite{dvf}& 324.15 & 68.61 &55.47  & 23.23   & 34.22 & 89.64 & 92.11 & 7.73   \\
DYAN~\cite{dyan} & 130.12 & 78.96 &70.41 & 13.09  & 21.43 & N/A & N/A & N/A          \\
OPT~\cite{wu2022optimizing} & 165312.80 &  83.26 & 73.85& 11.40 & 18.21 & 45716.20 & 96.75 & 3.59          \\
\midrule

DMVFN~(w/o r) & 19.39 & \textbf{84.81} & \textbf{75.05} &\textbf{ 9.41} & \textbf{16.24 }& 5.36 & \textbf{97.24} & \textbf{3.30}  \\ 
DMVFN & \textbf{9.96} & 83.97 & 74.81 & 9.96& 17.28 & \textbf{2.77} & 97.01 & 3.69   \\ 
\bottomrule
\end{tabular}
}
\label{table:davis-vimeo}
\end{table*}

\subsection{Implementation Details}
\label{sec:implement}

\noindent
\textbf{Loss function}. Our training loss ${L}_{total}$ is the sum of the reconstruction losses of outputs of each block ${I}^{i}_{t+1}$:
\begin{equation}\label{eq2}
     {L}_{total} = \sum_{i=1}^n \gamma^{n-i} d(\tilde{I}^{i}_{t+1}, I_{t+1}),
\end{equation}
where $d$ is the $\ell_1$ loss calculated on the Laplacian pyramid representations~\cite{laplacian} extracted from each pair of images. And we set $\gamma=0.8$ in our experiments following~\cite{raft}.

\begin{figure}[th]
	\centering
	\includegraphics[width=0.94\linewidth]{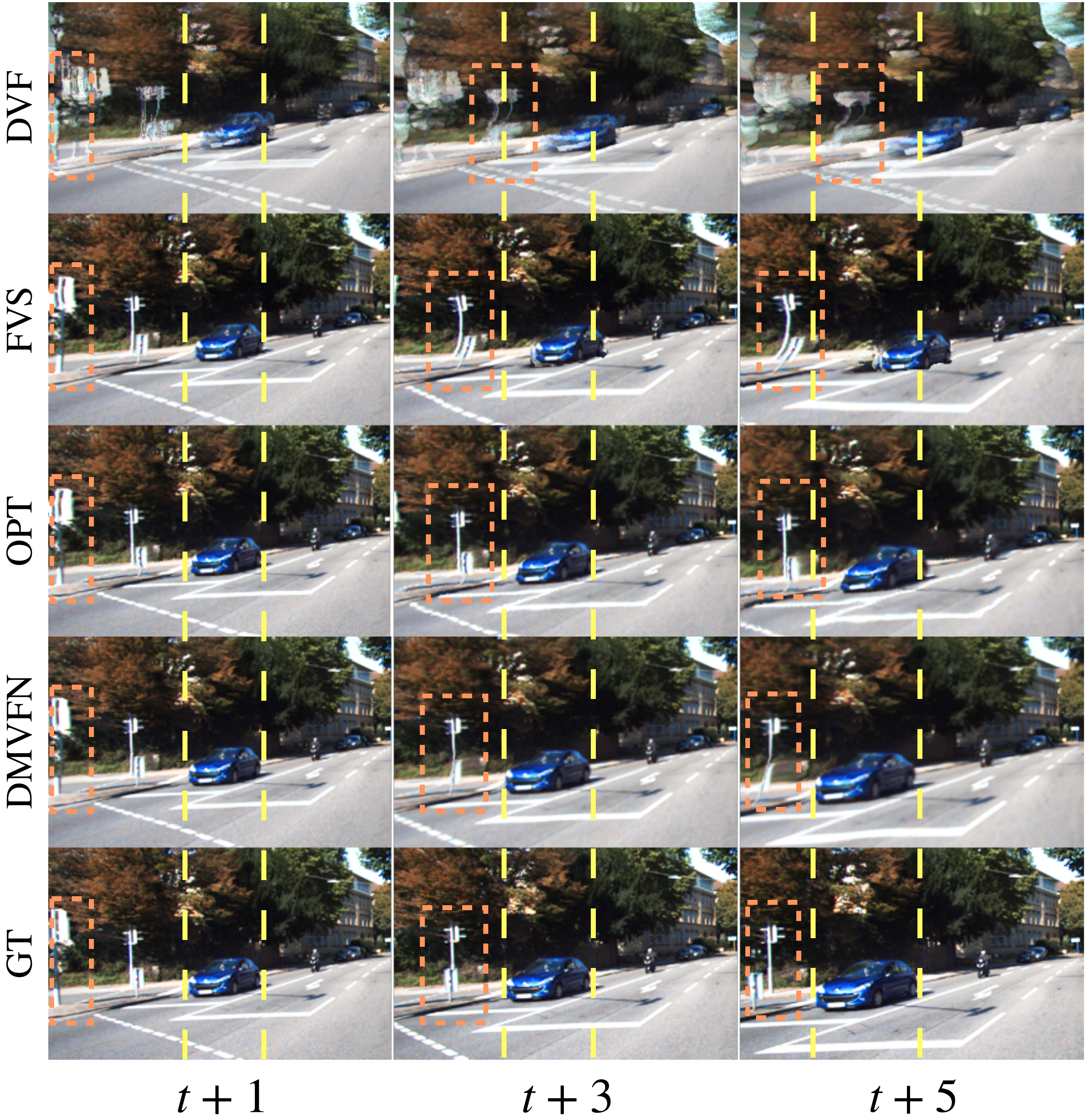}
	\caption{\textbf{Prediction comparison on KITTI}. The yellow line is aligned with the car in the ground truth. The results show that previous methods (DVF~\cite{dvf}, FVS~\cite{fvs}, and OPT~\cite{wu2022optimizing}) cannot accurately predict the car's location in the long-term prediction. The motion predicted by our DMVFN is the most similar to the ground truth, while the errors of other methods grow larger with time. The fences predicted by DMVFN remain vertical when moving.}
	\label{fig:kitti}
\end{figure}

\paragraph{Training strategy.} Our DMVFN is trained on $224\times224$ image patches. The batch size is set as $64$. 
We employ the AdamW optimizer~\cite{adam,loshchilov2018fixing} with a weight decay of $10^{-4}$.
We use a cosine annealing strategy to reduce the learning rate from $10^{-4}$ to $10^{-5}$.
Our model is trained on four 2080Ti GPUs for $300$ epochs, which takes about $35$ hours.

	\section{Experiments}
\begin{figure*}[th]
	\centering
	\includegraphics[width=0.98\linewidth]{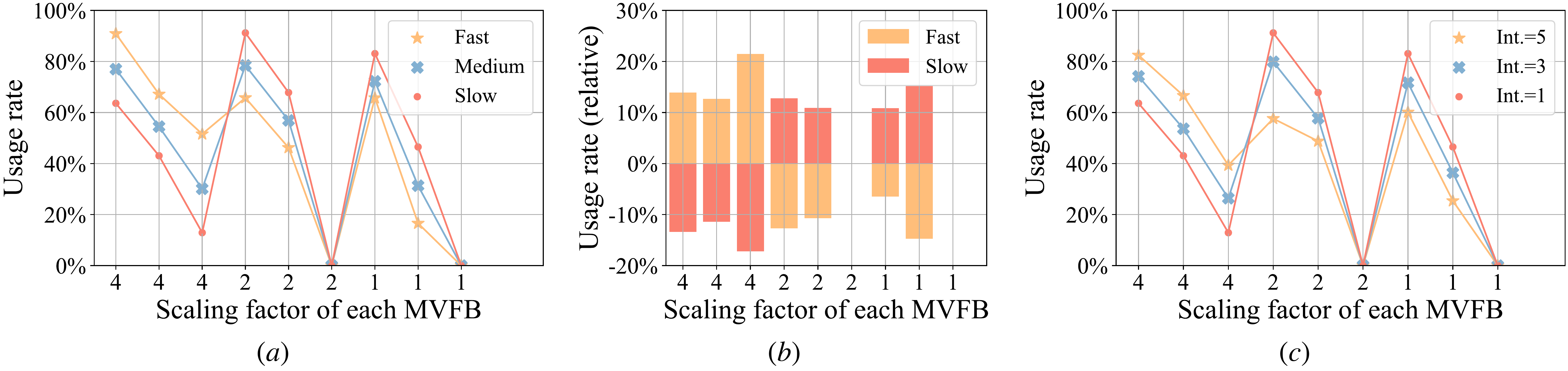}
	\caption{\textbf{$(a)$: Average usage rate on videos with different motion magnitudes}. ``Fast'': tested on Vimeo-Fast. ``Medium'': tested on Vimeo-Medium. ``Slow'': tested on Vimeo-Slow.
    \textbf{$(b)$: Difference between ``Fast''/``Slow'' and ``Medium'' of $(a)$.}
 \textbf{$(c)$: Averaged usage rate on different time intervals between two input frames from Vimeo-Slow}. ``Int.'': time interval.}
	\label{fig:ablation1}
\end{figure*}

 \begin{figure}[th]
 	\centering
 	\includegraphics[width=0.86\linewidth]{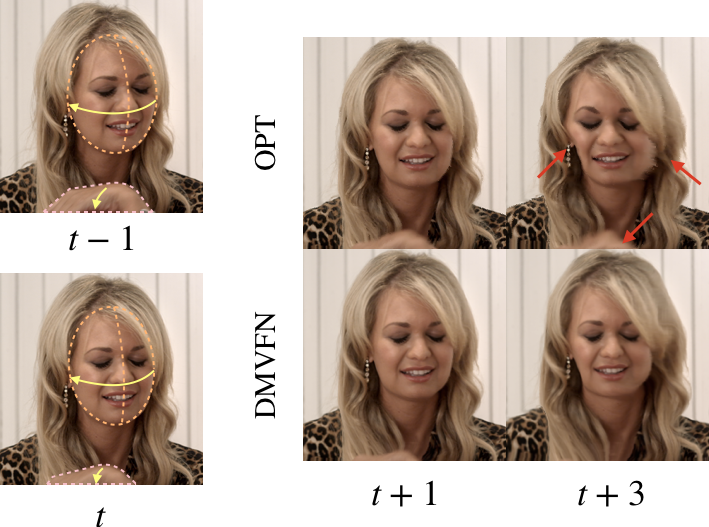}
 	\caption{\textbf{Visual effect comparison in the Vimeo-Test~\cite{vimeo} dataset.} our DMVFN faithfully reproduces the motion of the hand and the head with less distortion and artifacts.}
 	\label{fig:Vimeo}
    \vspace{-1em}
 \end{figure}

\subsection{Dataset and Metric}
\label{sec:dataset}
\paragraph{Dataset.} We use several datasets in the experiments:

\vspace{0.4em}
\noindent \textbf{Cityscapes} dataset~\cite{cityscapes} contains 3,475 driving videos with resolution of $2048 \times 1024$. We use 2,945 videos for training (Cityscapes-Train) and 500 videos in Cityscapes dataset~\cite{cityscapes} for testing (Cityscapes-Test).

\vspace{0.4em}
\noindent \textbf{KITTI} dataset~\cite{kitti} contains 28 driving videos with resolution of $375\times 1242$. 24 videos in KITTI dataset are used for training (KITTI-Train) and the remaining four videos in KITTI dataset are used for testing (KITTI-Test).

\vspace{0.4em}
\noindent \textbf{UCF101}~\cite{ucf101} dataset contains $13,320$ videos under 101 different action categories with resolution of $240\times 320$. We only use the training subset of UCF101~\cite{ucf101}.

\vspace{0.4em}
\noindent \textbf{Vimeo90K}~\cite{vimeo} dataset has $51,312$ triplets for training, where each triplet contains three consecutive video frames with resolution of $256\times448$.
There are $3,782$ triplets in the Vimeo90K testing set. We denote the training and testing subsets as Vimeo-Train and Vimeo-Test, respectively.

\vspace{0.4em}
\noindent
\textbf{DAVIS17~\cite{davis}} has videos with resolution around $854\times 480$. We use the DAVIS17-Val containing 30 videos as test set.

\noindent
\textbf{Configurations.} We have four experimental configurations following previous works~\cite{dyan,dvf,wu2022optimizing}:

\begin{itemize}
\itemsep-0.03in
\item Cityscapes-Train$\rightarrow$Cityscapes-Test
\item KITTI-Train$\rightarrow$KITTI-Test
\item UCF101$\rightarrow$DAVIS17-Val
\item UCF101$\rightarrow$Vimeo-Test
\end{itemize}
Here, the left and right sides of the arrow represent the training set and the test set, respectively. For a fair comparison with other methods that are not tailored for high resolution videos, we follow the setting in~\cite{fvs} and resize the images in Cityscapes~\cite{cityscapes} to $1024\times 512$ and images in KITTI~\cite{kitti} to $256\times 832$, respectively. During inference of Cityscapes~\cite{cityscapes} and KITTI~\cite{kitti}, we predict the next five frames. We predict the next three frames for DAVIS17-Val~\cite{davis} and next one frame for Vimeo-Test~\cite{vimeo}, respectively. Note that OPT~\cite{wu2022optimizing} is an optimization-based approach and uses pre-trained RAFT~\cite{raft} and RIFE~\cite{rife} models. RIFE~\cite{rife} and RAFT~\cite{raft} are trained on the Vimeo-Train dataset~\cite{vimeo} and the Flying Chairs dataset~\cite{flownet}, respectively.

\noindent
\textbf{Evaluation metrics}. Following previous works~\cite{wu2022optimizing}, we use Multi-Scale Structural Similarity Index Measure (MS-SSIM)~\cite{msssim} and a perceptual metric LPIPS~\cite{lpips} for quantitative evaluation. To measure the model complexity, we calculate the GFLOPs.

\subsection{Comparison to State-of-the-Arts}
\label{sec:comparison}
We compare our DMVFN with state-of-the-art video prediction methods. These methods fall into two categories: the methods requiring only RGB images as input (\eg, PredNet~\cite{prednet}, MCNET~\cite{mcnet}, DVF~\cite{dvf}, CorrWise~\cite{corrwise}, OPT~\cite{wu2022optimizing}) and the methods requiring extra information as input (\eg, Vid2vid~\cite{vid2vid}, Seg2vid~\cite{seg2vid}, FVS~\cite{fvs}, SADM~\cite{sadm}). 

\noindent
\textbf{Quantitative results}. The quantitative results are reported in Table~\ref{table:cityscapes} and Table~\ref{table:davis-vimeo}. When calculating the GFLOPs of OPT~\cite{wu2022optimizing}, the number of iterations is set as $3,000$. In terms of MS-SSIM and LPIPS, our DMVFN achieves much better results than the other methods in both short-term and long-term video prediction tasks. The GFLOPs of our DMVFN is considerably smaller than the comparison methods. These results show the proposed routing strategy reduces almost half the number of GFLOPs while maintaining comparable performance. Because the decrease of GFLOPs is not strictly linear with the actual latency~\cite{regnet}, we measure the running speed on TITAN 2080Ti. For predicting a 720P frame, DVF~\cite{dvf} spends $0.130$s on average, while our DMVFN only needs $0.023$s on average.

\begin{table}[!htb]
  \vspace{-2mm}
  \centering
  \caption{\textbf{Comparison between DMVFN and STRPM.%
  }}
  \vspace{-1mm}
  \label{tab:ucf}
   \resizebox{0.98\linewidth}{!}{ \begin{tabular}{lcccc}
    \toprule
    \multirow{3}{*}{Method}&\multicolumn{2}{c}{UCF Sports}&\multicolumn{2}{c}{Human3.6M}\cr
    \cmidrule(lr){2-3}\cmidrule(lr){4-5}
    &$t+1$&$t+6$&$t+1$&$t+4$\cr
    \cmidrule(lr){2-2} \cmidrule(lr){3-3}\cmidrule(lr){4-4} \cmidrule(lr){5-5}
    &PSNR$\uparrow$ / LPIPS$\downarrow$&PSNR$\uparrow$ / LPIPS$\downarrow$&PSNR$\uparrow$ / LPIPS$\downarrow$&PSNR$\uparrow$ / LPIPS$\downarrow$\cr
    \midrule
    STRPM                                                   &28.54 / 20.69  &20.59 / 41.11    &33.32 / 9.74  &29.01 / 10.44\cr
    DMVFN & \textbf{30.05} / \textbf{10.24}  &\textbf{22.67} / \textbf{22.50}  &\textbf{35.07} / \textbf{7.48}  &\textbf{29.56} / \textbf{9.74}\cr
    \bottomrule
    \end{tabular}}
    \vspace{-0.2em}
\end{table}

\noindent
\textbf{More comparison}. The quantitative results compared with STRPM~\cite{strpm} are reported in Table~\ref{tab:ucf}. We train our DMVFN in UCFSports and Human3.6M datasets following the setting in~\cite{strpm}. We also measure the average running speed on TITAN 2080Ti. To predict a $1024\times 1024$ frame, our DMVFN is averagely 4.06$\times$ faster than STRPM~\cite{strpm}.

\noindent
\textbf{Qualitative results} on different datasets are shown in Figure~\ref{fig:davis17}, Figure~\ref{fig:kitti} and Figure~\ref{fig:Vimeo}. As we can see, the frames predicted by our DMVFN exhibit better temporal continuity and are more consistent with the ground truth than those by the other methods. Our DMVFN is able to predict correct motion while preserving the shape and texture of objects.

\subsection{Ablation Study}
\label{sec:ablation}
Here, we perform extensive ablation studies to further study the effectiveness of components in our DMVFN. The experiments are performed on the Cityscapes~\cite{cityscapes} and KITTI~\cite{kitti} datasets unless otherwise specified.

\noindent
\textbf{1) How effective is the proposed Routing Module}? As suggested in~\cite{xiang2020zooming,TMNet2021}, we divide the Vimeo-90K~\cite{vimeo} test set into three subsets: Vimeo-Fast, Vimeo-Medium, and Vimeo-Slow, which correspond to the motion range. To verify that our DMVFN can perceive motion scales and adaptively choose the proper sub-networks, we retrain our DMVFN on the Vimeo-Train~\cite{vimeo} using the same training strategy in~\S\ref{sec:implement}. We calculate the averaged usage rate of each MVFB on three test subsets. From Figures~\ref{fig:ablation1}~$(a)$ and \ref{fig:ablation1}~$(b)$, we observe that our DMVFN prefers to select MVFBs with large scale (\eg, 4x) for two frames with large motion. There are two MVFBs with clearly smaller selection probability. We believe this reflects the inductive bias of our DMVFN on different combinations of scaling factors.

To further verify that our DMVFN also perceives the size of the time interval, we test our DMVFN on the two frames with different time intervals (but still in the same video). We choose Vimeo-Slow as the test set, and set the time intervals as $1$, $3$, and $5$. The results are shown in Figure~\ref{fig:ablation1}~$(c)$. We observe that our DMVFN prefers large-scale blocks on long-interval inputs, and small-scale blocks on short-interval inputs. This verifies that our DMVFN can perceive temporal information and dynamically select different sub-networks to handle the input frames with different time intervals.

To further study how the MVFBs are selected, we select 103 video sequences (contain a high-speed moving car and a relatively static background) from the KITTI dataset, denoted as KITTI-A. As shown in Table~\ref{tab:ablation1_1}, on the KITTI-A dataset, our DMVFN prefers to choose MVFBs with large scaling factors to capture large movements. The flow estimation for static backgrounds is straightforward, while the large motion dominates the choice of our DMVFN.

\begin{table}[!htb]
\vspace{-2mm}
  \centering
  \caption{\textbf{Average usage rate ($10^{-2}$) of MVFBs in our DMVFN.}}
  \label{tab:ablation1_1}
   \resizebox{0.98\linewidth}{!}{ \begin{tabular}{lccccccccc}
    \toprule
    Scale & $4$ & $4$& $4$& $2$ & $2$ & $2$ & $1$ & $1$ & $1$ \cr
    \midrule                           
    KITTI-A & $80.95$ &$34.22$ & $26.70$ & $81.19$ & $73.91$ & $44.90$ & $55.34$ & $0.49$& $0$\cr 
    \bottomrule
    \end{tabular}}
\end{table}

\begin{table}[th]
\caption{\textbf{Routing Module based on STEBS is effective}. The evaluation metric is MS-SSIM ($\times10^{-2}$).}

\centering
\resizebox{\linewidth}{!}{
\begin{tabular}{ccccccccccc}
\toprule
\multirow{2}{*}[-0.28em]{Setting} & \multicolumn{3}{c}{Cityscapes} & \multicolumn{3}{c}{KITTI}\\
\cmidrule(l{7pt}r{7pt}){2-4} \cmidrule(l{7pt}r{7pt}){5-7} 
& $t+1$ & $t+3$ & $t+5$ & $t+1$  & $t+3$ & $t+5$ \\ 
\midrule
\midrule
Copy last frame & 76.95 & 68.82 & 64.45 & 58.31 & 48.99 & 44.16 \\
w/o routing  & 95.29 & 87.91& 81.48 & 88.06 & 76.53 & 68.29 \\
Random & 91.97 & 82.11 & 70.05 & 81.31 & 69.89 & 62.42 \\
Gumbel Softmax & 95.05 & 87.57 & 79.54 & 87.42 & 75.56& 65.83   \\%
STEBS & \textbf{95.73}  & \textbf{89.24} & \textbf{83.45} &  \textbf{88.53} & \textbf{78.01}& \textbf{70.52} \\ %
\bottomrule
\end{tabular}
}
\label{table:gumbel}
\end{table}

\noindent
\textbf{2) How to design the Routing Module}? A trivial solution is to process the routing probability $p$ with Gumbel Softmax. The comparison results of our DMVFNs with different differentiable routing methods are summarized in Table~\ref{table:gumbel}. Our DMVFN with STEBS outperforms the DMVFN variant with Gumbel Softmax on MS-SSIM, especially for long-term prediction. The DMVFN variant with Gumbel Softmax usually degenerates to a fixed and static structure. We also compare with the DMVFN randomly selecting each MVFB with probability $0.5$~(denoted as ``Random'') and that without routing module (denoted as ``w/o routing'').

\begin{table}[th]
\caption{\textbf{Results of our DMVFN with different scaling factor settings}. The evaluation metric is MS-SSIM ($\times10^{-2}$).}
\centering
\resizebox{\linewidth}{!}{
\begin{tabular}{ccccccccc}
\toprule
Setting & \multicolumn{3}{c}{Cityscapes} & \multicolumn{3}{c}{KITTI}\\
\cmidrule(l{7pt}r{7pt}){2-4} \cmidrule(l{7pt}r{7pt}){5-7}
in DMVFN  & $t+1$ & $t+3$ & $t+5$  & $t+1$  & $t+3$ & $t+5$  \\ 
\midrule
\midrule
 $[1]$ & 94.70 & 87.26 & 80.93 & 87.64 & 76.71 & 68.76 \\
 $[2,1]$ & 95.30 & 87.93 & 82.02 & 87.97 & 77.23 & 69.58 \\
 $[4,2,1]$ & \textbf{95.73}  & \textbf{89.24} & \textbf{83.45} & \textbf{88.53} & \textbf{78.01} & \textbf{70.52} \\

\bottomrule
\end{tabular}
}
\label{table:blocks}
\end{table}

\noindent
\textbf{3) How to set the scaling factors}?\ We evaluate our DMVFN with different scaling factors.\ We use three non-increasing factor sequences of ``$[1,1,1,1,1,1,1,1,1]$'', ``$[2,2,2,2,2,1,1,1,1]$'' and ``$[4,4,4,2,2,2,1,1,1]$'', denoted as ``$[1]$", ``$[2,1]$" and ``$[4,2,1]$", respectively. The results are listed in Table~\ref{table:blocks}. Our DMVFN with ``$[4,2,1]$'' performs better than that with ``$[2,1]$'' and ``$[1]$''. The gap is more obvious on longer-term future frames.

\begin{table}[th]
\caption{\textbf{Spatial path is effective in our DMVFN}. The evaluation metric is MS-SSIM ($\times10^{-2}$).}

\centering
\resizebox{\linewidth}{!}{
\begin{tabular}{ccccccccc}
\toprule
\multirow{2}{*}[-0.28em]{Setting} & \multicolumn{3}{c}{Cityscapes} & \multicolumn{3}{c}{KITTI}\\
\cmidrule(l{7pt}r{7pt}){2-4} \cmidrule(l{7pt}r{7pt}){5-7} & $t+1$ & $t+3$ & $t+5$  & $t+1$  & $t+3$ & $t+5$  \\ 
\midrule
\midrule
w/o r, w/o path & 94.99 & 87.59 & 80.98 & 87.75 & 76.22 & 67.86 \\
w/o r & 95.29 & 87.91 & 81.48 & 88.06 & 76.53 & 68.29 \\
w/o path &  95.55 & 88.89 & 83.03 & 88.29 & 77.53 & 69.86 \\%
DMVFN & \textbf{95.73}  & \textbf{89.24} & \textbf{83.45} & \textbf{88.53} & \textbf{78.01} & \textbf{70.52} \\

\bottomrule
\end{tabular}
}
\label{table:path}
\end{table}

\noindent
\textbf{4) How effective is the spatial path}? To verify the effectiveness of the spatial path in our DMVFN, we compare it with the DMVFN without spatial path (denoted as ``w/o path''). The results listed in Table~\ref{table:path} show our DMVFN enjoys better performance with the spatial path, no matter with or without the routing module (denoted as ``w/o r'').

	\section{Conclusion}
 
In this work, we developed an efficient Dynamic Multi-scale Voxel Flow Network (DMVFN) that excels previous video prediction methods on dealing with complex motions of different scales. With the proposed routing module, our DMVFN adaptively activates different sub-networks based on the input frames, improving the prediction performance while reducing the computation costs. Experiments on diverse benchmark datasets demonstrated that our DMVFN achieves \sota performance with greatly reduced computation burden. We believe our DMVFN can provide general insights for long-term prediction, video frame synthesis, and representation learning~\cite{hafner2023mastering,ha2018world}. We hope our DMVFN will inspire further research in light-weight video processing and make video prediction more accessible for downstream tasks such as CODEC for streaming video.

Our DMVFN can be improved at several aspects. Firstly, iteratively predicting future frames suffers from accumulate errors.
This issue may be addressed by further bringing explicit temporal modeling~\cite{qvi,TMNet2021,ifrnet,rife} to our DMVFN.
Secondly, our DMVFN simply selects the nodes in a chain network topology, which can be improved by exploring more complex topology. For example, our routing module can be extended to automatically determine the scaling factors for parallel branches~\cite{darts}.
Thirdly, forecast uncertainty modeling is more of an extrapolation abiding to past flow information, especially considering bifurcation, which exceeds the current capability of our DMVFN. We believe that research on long-term forecast uncertainty may uncover deeper interplay with dynamic modeling methods~\cite{ha2018world,akan2021slamp}.

\noindent
\textbf{Acknowledgements}. We sincerely thank Wen Heng for his exploration on neural architecture search at Megvii Research and Tianyuan Zhang for meaningful suggestions. This work is supported in part by the National Natural Science Foundation of China (No. 62002176 and 62176068).
        \clearpage
	{\small
		\bibliographystyle{ieee_fullname}    
		\bibliography{egbib}
	}
        \clearpage
        \onecolumn

\begin{center}\Large{\textbf{Supplemental File to ``A Dynamic Multi-Scale Voxel Flow Network for Video Prediction''}}
\end{center}

We provide 
more details of DMVFN for video prediction.
Specifically, we provide
\begin{itemize}
\itemsep-0.03in
\item societal impact in~\S\ref{sec:impact}.

\item visualization of voxel flow~\S\ref{sec:vf};

\item more ablation studies in~\S\ref{sec:more_ab};

\end{itemize}

\section{Societal Impact}
\label{sec:impact}
This work potentially benefits video prediction and dynamic neural network fields.\ The authors believe that this work has small potential negative impacts.

\section{Visualization of Voxel Flow}
\begin{figure*}[h]
\vspace{4mm}
	\centering
	\includegraphics[width=\textwidth]{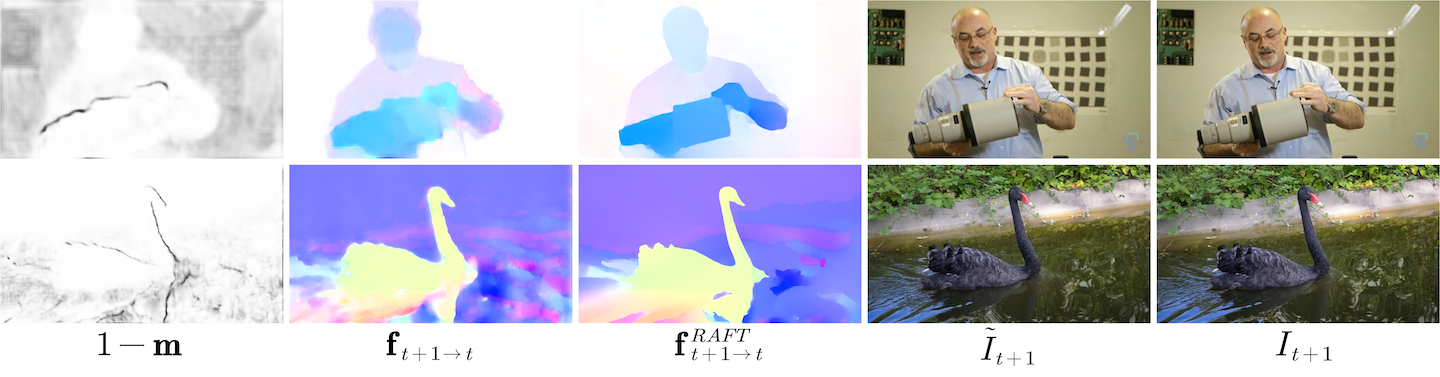}
 \caption{\textbf{Visualization} of the map $1-\textbf{m}$, the optical flow $\textbf{f}_{t+1 \rightarrow t}$, the optical flow by RAFT~\cite{raft} $\textbf{f}^{RAFT}_{t+1 \rightarrow t}$, the predicted frame $\tilde{I}_{t+1}$ and the “ground truth” $I_{t+1}$.}
        \label{fig:vf}
	\vspace{-2mm}
\end{figure*}

We visualize the voxel flow predicted by DMVFN in Figure~\ref{fig:vf}.
We use the optical flow generated by RAFT~\cite{raft} as a reference.
We observe that the optical flow $\textbf{f}_{t+1 \rightarrow t}$ and the map $1-\textbf{m}$ of most pixels are successfully predicted by DMVFN.
This demonstrates that our DMVFN can indeed accurately predict a voxel flow.
\label{sec:vf}

\section{More Ablation Study Results}
\label{sec:more_ab}

\noindent
\textbf{5) How does $\beta$ influence the performance of DMVFN during inference?} The $\beta$ is an important factor to control the model complexity and prediction capability during inference. Here, we adjust $\beta$ during the inference phase, as shown in Table~\ref{table:beta}. DMVFN with larger $\beta$ enjoys better MS-SSIM results but suffers from higher complexity.

\begin{table}[th]
\caption{\textbf{Results of DMVFN with different $\beta$}\ evaluated on KITTI benchmark~\cite{kitti}.}
\centering
\resizebox{0.66\linewidth}{!}{
\begin{tabular}{lccccccccc}
\toprule
Settings~($\beta=$)   & $0.3$ & $0.4$ & $0.5$ & $0.6$ & $0.7$ & $0.8$  \\
\midrule
GFLOPs & \textbf{2.62} & 3.88 & 5.15 & 5.94 & 6.21 & 6.40 \\
LPIPS & 16.47 & 12.91 & 10.74 & 10.26& 10.24 & \textbf{10.23} \\ 
MS-SSIM ($\times10^{-2}$) & 78.78 & 85.13 & 88.53 & 88.89 & 88.89 & 88.89\\
\bottomrule
\end{tabular}
}
\label{table:beta}
\end{table}

\begin{figure}[th]
 	\centering
 \includegraphics[width=0.55\linewidth]{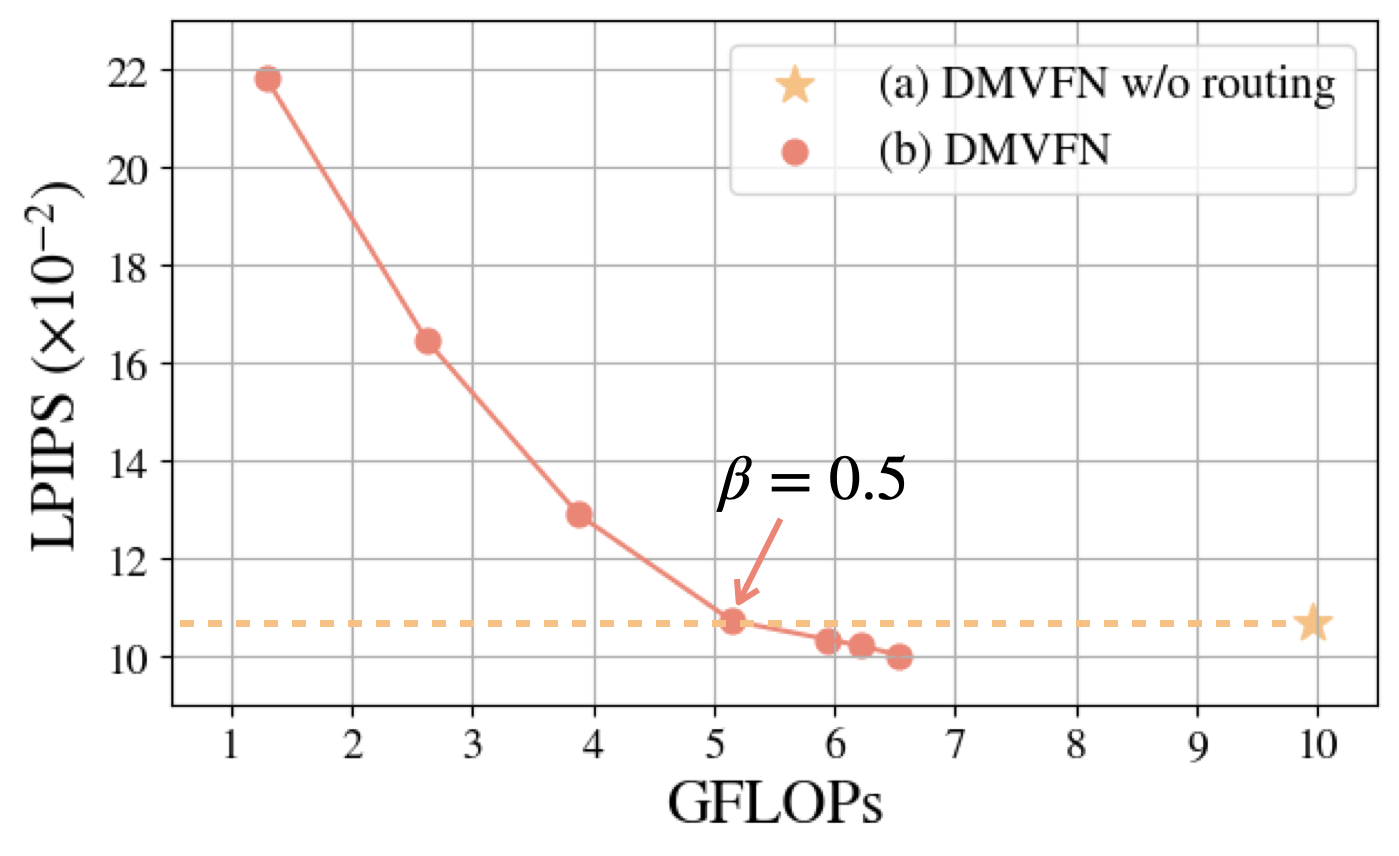}
  	\caption{\textbf{Control the complexity of DMVFN by adjusting $\beta$}. DMVFN saves half GFLOPs of comparable performance compared to DMVFN without routing.}
 	\label{fig:flops}
 \end{figure}

\noindent
\textbf{6) How to design the loss function?} To study this problem, we train our DMVFN and DMVFN (w/o routing) only optimizing the loss on output of the last block $\tilde{I}_{t+1}$~(denoted as ``single supervision"). The results listed in Table~\ref{table:ab6} show the advantages of our loss function $L_{total}$. $L_{total}$ is calculated on all intermediate results of DMVFN.
\begin{table*}[th]
\caption{\textbf{Results of DMVFN with different loss settings}. The evaluation metric is MS-SSIM ($\times10^{-2}$).}

\centering
\resizebox{\linewidth}{!}{
\begin{tabular}{cccccccccccc}
\toprule
\multirow{2}{*}[-0.28em]{Settings} & \multicolumn{3}{c}{Cityscapes} & \multicolumn{3}{c}{KITTI} & \multicolumn{2}{c}{Davis17-Val} & \multicolumn{1}{c}{Vimeo-Test}\\
\cmidrule(l{7pt}r{7pt}){2-4} \cmidrule(l{7pt}r{7pt}){5-7} \cmidrule(l{7pt}r{7pt}){8-9} \cmidrule(l{7pt}r{7pt}){10-10} 
 & t+1 & t+3 & t+5  & t+1  & t+3 & t+5 & t+1 & t+3  & t+1  \\ 
\midrule
\midrule
w/o routing, single supervision  & 95.19 & 87.77 & 81.22 & 87.91 & 76.33 & 67.99 & 84.69 & 74.92 & 97.18 \\
w/o routing & 95.29 & 87.91& 81.48 & 88.06 & 76.53 & 68.29 & \textbf{84.81} & \textbf{75.05} & \textbf{97.24} \\
single supervision & 95.65 & 89.10 & 83.27 & 88.34 & 77.88 & 70.18 & 83.83 & 74.68 & 96.95 \\%
DMVFN & \textbf{95.73}  & \textbf{89.24} & \textbf{83.45} & \textbf{88.53} & \textbf{78.01} & \textbf{70.52} & 83.97 & 74.81 & 97.01\\

\bottomrule
\end{tabular}
}
\label{table:ab6}
\end{table*}

\begin{table*}[th]
\caption{\textbf{Routing Module based on STEBS is effective}. The evaluation metric is MS-SSIM ($\times10^{-2}$).}

\centering
\resizebox{\linewidth}{!}{
\begin{tabular}{cccccccccccc}
\toprule
\multirow{2}{*}[-0.28em]{Settings} & \multicolumn{3}{c}{Cityscapes} & \multicolumn{3}{c}{KITTI} & \multicolumn{2}{c}{Davis-Val} & \multicolumn{1}{c}{Vimeo-Test}\\
\cmidrule(l{7pt}r{7pt}){2-4} \cmidrule(l{7pt}r{7pt}){5-7} \cmidrule(l{7pt}r{7pt}){8-9} \cmidrule(l{7pt}r{7pt}){10-10} 
 & t+1 & t+3 & t+5  & t+1  & t+3 & t+5 & t+1 & t+3  & t+1  \\ 
\midrule
\midrule
w/o routing  & 95.29 & 87.91& 81.48 & 88.06 & 76.53 & 68.29 & \textbf{84.81} & \textbf{75.05} & \textbf{97.24} \\
Random & 91.97 & 82.11 & 70.05 & 81.31 & 69.89 & 62.42 & 81.32 & 73.03 & 96.88 \\
Gumbel Softmax & 95.05 & 87.57 & 79.54 & 87.42 & 75.56& 65.83 & 83.64 & 74.43 & 96.98  \\%
STEBS & \textbf{95.73}  & \textbf{89.24} & \textbf{83.45} & \textbf{88.53} & \textbf{78.01} & \textbf{70.52} & 83.97 & 74.81 & 97.01\\

\bottomrule
\end{tabular}
}
\label{table:ab21}
\end{table*}

\noindent
\textbf{More details about our Ablation Study 2) in the main paper}.
In Table~\ref{table:ab21}, we summarize the quantitative results of three variants (``w/o routing'', ``Random'' and ``Gumbel Softmax'') on four datasets (i.e., Cityscapes~\cite{cityscapes}, KITTI~\cite{kitti}, Davis-Val~\cite{davis}, and Vimeo-Test~\cite{vimeo}). This demonstrates the effectiveness of our STEBS.

\noindent
\textbf{More details about our Ablation Study 3) in the main paper}.
In Table~\ref{table:ab31}, we summarize the quantitative results of DMVFN with different scaling factor settings, including:
\begin{itemize}
\itemsep-0.03in
\item ``[1]'': [1,1,1,1,1,1,1,1,1]
\item ``[2]'': [2,2,2,2,2,2,2,2,2]
\item ``[4]'': [4,4,4,4,4,4,4,4,4]
\item ``[1,2]'': [1,1,1,1,2,2,2,2,2]
\item ``[1,4]'': [1,1,1,1,4,4,4,4,4]
\item ``[2,1]'': [2,2,2,2,1,1,1,1,1]
\item ``[4,1]'': [4,4,4,4,1,1,1,1,1]
\item ``[1,2,4]'': [1,1,1,2,2,2,4,4,4]
\item ``[4,2,1]'': [4,4,4,2,2,2,1,1,1]
\end{itemize}
DMVFN [4,2,1] performs better than others, and the gap is more obvious for long-term future frames.

\begin{table*}[th]
\caption{\textbf{Results of DMVFN with different scaling factor settings}. The evaluation metric is MS-SSIM ($\times10^{-2}$).}

\centering
\resizebox{\linewidth}{!}{
\begin{tabular}{cccccccccccc}
\toprule
\multirow{2}{*}[-0.28em]{Settings} & \multicolumn{3}{c}{Cityscapes} & \multicolumn{3}{c}{KITTI} & \multicolumn{2}{c}{Davis-Val} & \multicolumn{1}{c}{Vimeo-Test}\\
\cmidrule(l{7pt}r{7pt}){2-4} \cmidrule(l{7pt}r{7pt}){5-7} \cmidrule(l{7pt}r{7pt}){8-9} \cmidrule(l{7pt}r{7pt}){10-10} 
 & t+1 & t+3 & t+5  & t+1  & t+3 & t+5 & t+1 & t+3  & t+1  \\ 
\midrule
\midrule
DMVFN [1] & 94.70 & 87.26 & 80.93 & 87.64 & 76.71 & 68.76 & 81.75 & 71.73 & 96.04 \\
DMVFN [2] & 95.51 & 87.76 & 81.30 &  87.06 &  76.90 & 69.05  &  81.77 & 72.58  & 96.07  \\
DMVFN [4] &  94.32 & 87.50 & 81.36  &  84.35 & 75.34  & 68.67 & 81.02 & 72.16 &  95.99 \\
DMVFN [1, 2] &  94.13 & 86.58  & 80.55 &  87.85 & 76.92  &  69.36 & 82.96  & 73.55  & 96.70  \\
DMVFN [1, 4] & 94.56 & 86.50 & 80.69 & 85.46  & 76.03  &  68.99 & 81.38  & 71.98  & 96.02  \\
DMVFN [2, 1] & 95.30 & 87.93 & 82.02 & 87.97 & 77.23 & 69.58 & 83.03 & 72.54 & 96.61 \\
DMVFN [4, 1] & 95.59 & 88.41 & 83.02 & 88.16  & 77.39  & 69.95  & 83.64  &  74.35 & 96.95  \\
DMVFN [1, 2, 4] & 94.20 & 86.56 & 80.81  &  87.77 & 76.89  & 69.72  &  82.72 &  73.66 &  96.76 \\
DMVFN [4, 2, 1] & \textbf{95.73}  & \textbf{89.24} & \textbf{83.45} & \textbf{88.53} & \textbf{78.01} & \textbf{70.52} & \textbf{83.97} & \textbf{74.81} & \textbf{97.01}\\

\bottomrule
\end{tabular}
}
\label{table:ab31}
\end{table*}

\noindent
\textbf{More details about our Ablation Study 4) in the main paper}.
In Table~\ref{table:ab41}, we summarize the quantitative results of three variants (``w/o r, w/o path'', ``w/o r'' and ``w/o path'') on four datasets (i.e., Cityscapes~\cite{cityscapes}, KITTI~\cite{kitti}, Davis-Val~\cite{davis}, and Vimeo-Test~\cite{vimeo}).
\begin{table*}[th]
\caption{\textbf{Spatial path is effective in DMVFN}. The evaluation metric is MS-SSIM ($\times10^{-2}$).}

\centering
\resizebox{\linewidth}{!}{
\begin{tabular}{cccccccccccc}
\toprule
\multirow{2}{*}[-0.28em]{Settings} & \multicolumn{3}{c}{Cityscapes} & \multicolumn{3}{c}{KITTI} & \multicolumn{2}{c}{Davis-Val} & \multicolumn{1}{c}{Vimeo-Test}\\
\cmidrule(l{7pt}r{7pt}){2-4} \cmidrule(l{7pt}r{7pt}){5-7} \cmidrule(l{7pt}r{7pt}){8-9} \cmidrule(l{7pt}r{7pt}){10-10} 
 & t+1 & t+3 & t+5  & t+1  & t+3 & t+5 & t+1 & t+3  & t+1  \\ 
\midrule
\midrule
w/o r, w/o path & 94.99 & 87.59 & 80.98 & 87.75 & 76.22 & 67.86   &  84.45 & 74.78  & 97.05\\
w/o r & 95.29 & 87.91 & 81.48 & 88.06 & 76.53 & 68.29  & \textbf{84.81} & \textbf{75.05} & \textbf{97.24} \\
w/o path &  95.55 & 88.89 & 83.03 & 88.29 & 77.53 & 69.86  & 83.75  & 74.51  & 96.89  \\
DMVFN & \textbf{95.73}  & \textbf{89.24} & \textbf{83.45} & \textbf{88.53} & \textbf{78.01} & \textbf{70.52} & 83.97 & 74.81 & 97.01\\

\bottomrule
\end{tabular}
}
\label{table:ab41}
\end{table*}

\noindent
\textbf{More details about our Ablation Study 5) in the main paper}.
In Table~\ref{table:ab51}, we summarize the quantitative results of different $\beta$ during inference on four datasets (i.e., Cityscapes~\cite{cityscapes}, KITTI~\cite{kitti}, Davis-Val~\cite{davis}, and Vimeo-Test~\cite{vimeo}).

\begin{table*}[th]
\caption{\textbf{Results of DMVFN with different $\beta$}\ evaluated on Cityscapes benchmark~\cite{cityscapes} and Vimeo-Test benchmark~\cite{vimeo}.}
\centering
\resizebox{\linewidth}{!}{
\begin{tabular}{ccccccccccccc}
\toprule
Settings & \multicolumn{6}{c}{Cityscapes} & \multicolumn{6}{c}{Vimeo-Test}\\
\cmidrule(l{7pt}r{7pt}){2-7} \cmidrule(l{7pt}r{7pt}){8-13}
$\beta=$   & $0.3$ & $0.4$ & $0.5$ & $0.6$ & $0.7$ & $0.8$ & $0.3$ & $0.4$ & $0.5$ & $0.6$ & $0.7$ & $0.8$ \\ %
\midrule
GFLOPs & \textbf{6.56} & 9.81 & 12.71 & 15.30 & 16.23 & 17.82 & \textbf{1.38} & 2.08 & 2.77 & 3.40 & 3.74 & 3.92 \\ 
LPIPS & 8.88 &  7.06 &  5.58 & 5.20 & 5.15 & \textbf{5.12} & 5.18 & 4.18 &   3.69 & 3.48 & 3.42 & \textbf{3.40} \\   
MS-SSIM ($\times10^{-2}$) &90.48  & 93.54 & 95.73 & 96.03 & 96.07 & \textbf{96.12} & 93.61 & 96.13 & 97.01 & 97.19 & \textbf{97.20} & \textbf{97.20}\\
\bottomrule
\end{tabular}
}
\label{table:ab51}
\end{table*}

	\clearpage
\end{document}